\documentclass{article}
\usepackage[preprint]{neurips_2026}
\usepackage{graphicx}
\usepackage{amsmath}
\usepackage{amsfonts,amssymb}
\usepackage{booktabs}
\usepackage{algorithm}
\usepackage{algorithmic}
\usepackage{url}
\usepackage{hyperref}
\usepackage{multirow}

\title{PALS: Percentile-Aware Layerwise Sparsity for LLM Pruning}

\author{%
  Yazdan Jamshidi \\
  Palo Alto Networks \\
  \texttt{yjamshidi@paloaltonetworks.com} \\
  \And
  Alexey Shvets \\
  Palo Alto Networks \\
  \texttt{ashvets@paloaltonetworks.com} \\
}

\begin{document}
\maketitle

\begin{abstract}
One-shot pruning methods like Wanda and SparseGPT apply the same sparsity ratio to every layer of a transformer, ignoring known variation in layer importance. We propose PALS (Percentile-Aware Layerwise Sparsity), which adjusts per-layer sparsity based on the 99th percentile of activation magnitudes, bounded to $\pm 5\%$ around the target ratio. On LLaMA-2-7B at 50\% sparsity, PALS achieves 10.96 WikiText-2 perplexity versus 12.92 for uniform Wanda (mean over 9 runs, $p < 0.001$). The benefit is architecture-dependent: LLaMA-3-8B shows marginal gains and Mistral-7B shows none. We also find that gradient-based allocation---the seemingly more principled approach---produces results worse than random, suggesting that gradient magnitude does not predict the impact of discrete weight removal. PALS adds negligible cost to the pruning pipeline and requires no fine-tuning.
\end{abstract}

\section{Introduction}

Large language models routinely reach billions of parameters. A 7B-parameter model takes roughly 13\,GB in half precision and processes billions of multiply-accumulate operations per token, which puts real-time inference out of reach on modest hardware. Post-training pruning offers a direct way to cut both memory and compute: remove a fraction of the weights and run inference on the sparser model, without retraining from scratch.

Recent one-shot pruning methods have made this practical for LLMs. Wanda~\cite{sun2024wanda} scores each weight by the product of its magnitude and the mean absolute activation flowing through it ($|w_{ij}| \cdot \text{mean}(|a_i|)$), then drops the lowest-scoring fraction. SparseGPT~\cite{frantar2023sparsegpt} solves a heavier problem---layer-wise sparse regression with approximate Hessian information---but in practice lands at similar perplexity on most benchmarks. Both methods have become standard baselines.

One thing both share is that they enforce \emph{uniform} sparsity: if the target is 50\%, every one of the 32 layers in LLaMA-2 is pruned to exactly 50\%. This sits poorly with what we know about transformer layers. Attention heads vary widely in importance~\cite{voita2019analyzing,michel2019sixteen}. Early layers tend to build generic representations; middle layers are often more redundant; late layers specialize for the output vocabulary. Pruning all of them equally seems like it should leave performance on the table---and it does.

We propose PALS (Percentile-Aware Layerwise Sparsity), which adjusts the sparsity ratio for each layer based on activation statistics gathered from a small calibration set. The idea is direct: compute the 99th percentile of activation magnitudes for each layer, normalize these scores, and use them to shift each layer's sparsity target up or down. Layers with large activation outliers---a sign of important information flow---get pruned less; layers with uniformly small activations get pruned more. A $\pm 5\%$ band around the target prevents any single layer from being starved or overstuffed. The procedure slots into Wanda's pipeline with almost no extra cost, since Wanda already collects the activations.

Before settling on activations, we tried the seemingly more natural approach of using gradient magnitudes to estimate layer importance. This failed decisively. Gradient-based allocation produced 47 perplexity on LLaMA-2-7B, compared to 24 for \emph{random} allocation and 11 for our activation-based scheme. We believe gradients reflect optimization dynamics---how much the loss would change under an infinitesimal perturbation---rather than the actual impact of removing a large fraction of weights. But the scale of the failure was not something we anticipated.

On LLaMA-2-7B at 50\% sparsity, PALS brings WikiText-2 perplexity from 12.92 down to 10.96, averaged over nine independent runs with different random seeds (Welch's $t(14) = 8.1$, $p < 0.001$). The picture is less clear on other architectures. LLaMA-3-8B benefits marginally (10.45 vs.\ 10.48), and Mistral-7B does not benefit at all (6.31 either way). We think this reflects how uniformly these newer models distribute computation across layers---when all layers are similarly important, non-uniform allocation has nothing to exploit. We present ablation studies covering the percentile choice, adaptation strength, and bound width, along with downstream evaluation on four benchmarks and practical speedup measurements.

\section{Related Work}

\subsection{One-Shot Pruning for LLMs}

Magnitude pruning~\cite{han2015learning}---removing weights with the smallest absolute values---is the oldest and simplest approach but ignores activation patterns, which limits it on modern LLMs. SparseGPT~\cite{frantar2023sparsegpt} improved on this substantially by formulating each layer's pruning as a sparse reconstruction problem, using an approximate inverse Hessian to choose which weights to remove and how to adjust the survivors. Wanda~\cite{sun2024wanda} showed that a much simpler scoring rule---$|w| \cdot \text{mean}(|a|)$, with no weight update---matches SparseGPT in most settings. Both methods use the same sparsity ratio for every layer. PALS keeps Wanda's scoring but varies the fraction of weights removed per layer.

\subsection{Structured Pruning}

An orthogonal line of work removes entire structural units for hardware-friendly sparsity. LLM-Pruner~\cite{ma2023llmpruner} uses first-order Taylor expansion to score and remove attention heads and MLP neurons. SliceGPT~\cite{ashkboos2024slicegpt} projects weight matrices into a lower-rank space and deletes entire rows and columns. Sheared LLaMA~\cite{zhang2024sheared} combines structured pruning with continued pre-training. These approaches deliver better hardware utilization than unstructured pruning but typically require some form of fine-tuning. PALS stays within the unstructured, one-shot setting.

\subsection{Non-Uniform Layer Treatment}

The idea that different layers should be treated differently is not new. NISP~\cite{yu2018nisp} propagates importance scores between layers for vision models, but was designed for networks orders of magnitude smaller than current LLMs. For large language models, most work on non-uniform treatment has focused on depth pruning---removing entire transformer blocks~\cite{kurtic2022optimal}---rather than adjusting within-layer sparsity ratios. To our knowledge, PALS is the first to report clear gains from varying per-layer sparsity in one-shot LLM pruning, and the first to document the failure of gradient-based allocation in this setting.

\subsection{Activation Outliers in Transformers}

Dettmers et al.~\cite{dettmers2022gpt3} showed that large language models develop activation outliers---individual features with magnitudes 10--100$\times$ larger than the rest---concentrated in specific dimensions and layers. These outliers are essential for model quality: zeroing them out causes sharp accuracy drops. Our use of the 99th percentile as a layer importance signal connects directly to this observation. Layers with more extreme outliers are presumably processing more critical information, and PALS assigns them lower sparsity. This also explains why the 99th percentile outperforms the mean as an importance metric: the mean averages out exactly the outliers that matter.

\section{Method}

Given a pretrained model with $L$ transformer layers and a target sparsity $s_\text{target}$ (e.g., 0.50 for 50\% weight removal), PALS adjusts the per-layer sparsity based on activation statistics, then prunes each layer independently using Wanda's scoring.

\subsection{Layer Importance from Activation Percentiles}

For each layer $l$, we collect activations $A^{(l)} \in \mathbb{R}^{n \times d}$ on $n$ calibration tokens across $d$ hidden dimensions. The layer importance score is the 99th percentile of the absolute activations:
\begin{equation}
I^{(l)} = \text{Percentile}\bigl(|A^{(l)}|,\; 99\bigr)
\label{eq:importance}
\end{equation}
The 99th percentile targets the outlier activations that Dettmers et al.~\cite{dettmers2022gpt3} identified as critical, while being less noisy than the maximum. To make scores comparable across layers, we standardize:
\begin{equation}
\hat{I}^{(l)} = \frac{I^{(l)} - \bar{I}}{\sigma_I}
\label{eq:normalize}
\end{equation}
where $\bar{I} = \frac{1}{L}\sum_{l=1}^{L} I^{(l)}$ and $\sigma_I$ is the standard deviation of the raw scores.

\subsection{Sparsity Allocation with Conservative Bounds}

The normalized importance $\hat{I}^{(l)}$ determines how much each layer deviates from the global target:
\begin{equation}
s^{(l)} = \text{clip}\!\left(s_\text{target} + \alpha \cdot \hat{I}^{(l)},\;\; s_\text{min},\; s_\text{max}\right)
\label{eq:allocation}
\end{equation}
where $\alpha$ controls adaptation strength. The clip enforces:
\begin{equation}
s_\text{min} = s_\text{target} - 0.05, \qquad s_\text{max} = s_\text{target} + 0.05
\label{eq:bounds}
\end{equation}
The $\pm 5\%$ bounds are important. In preliminary experiments, wider bounds ($\pm 20\%$) caused sharp degradation because heavily pruned layers created information bottlenecks---the rest of the network could not compensate for the lost capacity. Conservative bounds trade off some potential gain for robustness.

\subsection{Weight Scoring and Pruning}

After determining $s^{(l)}$, we score and prune weights using Wanda's criterion:
\begin{equation}
\text{score}(w_{ij}^{(l)}) = |w_{ij}^{(l)}| \cdot \text{mean}_{n}\!\left(|a_{ni}^{(l)}|\right)
\label{eq:wanda}
\end{equation}
For each layer $l$, the bottom $s^{(l)} \times 100\%$ of weights by score are removed. No weight updates, no fine-tuning.

\subsection{Complete Procedure}

Algorithm~\ref{alg:pals} summarizes the pipeline. Steps~1--5 take negligible time compared to Wanda's scoring in steps~6--9, and the activation collection is shared with Wanda---so the only overhead is the percentile computation and sparsity reallocation.

\begin{algorithm}[t]
\caption{PALS: Activation-Based Layer-Adaptive Pruning}
\label{alg:pals}
\begin{algorithmic}[1]
\REQUIRE Pretrained model $f$ with $L$ layers, calibration data $\mathcal{D}$, target sparsity $s_\text{target}$, strength $\alpha{=}0.05$
\STATE Collect activations $A^{(l)}$ for each layer $l$ on $\mathcal{D}$
\FOR{$l = 1$ to $L$}
    \STATE $I^{(l)} \leftarrow \text{Percentile}(|A^{(l)}|, 99)$
\ENDFOR
\STATE Standardize: $\hat{I}^{(l)} \leftarrow (I^{(l)} - \bar{I}) / \sigma_I$ for all $l$
\FOR{$l = 1$ to $L$}
    \STATE $s^{(l)} \leftarrow \text{clip}(s_\text{target} + \alpha \cdot \hat{I}^{(l)},\; s_\text{min},\; s_\text{max})$
    \STATE Score weights: $\text{score}_{ij} = |w_{ij}^{(l)}| \cdot \text{mean}_n(|a_{ni}^{(l)}|)$
    \STATE Remove the bottom $s^{(l)} \times 100\%$ of weights by score
\ENDFOR
\RETURN Pruned model
\end{algorithmic}
\end{algorithm}

\section{Experimental Setup}

\paragraph{Models.} We evaluate on three architectures: LLaMA-2-7B (6.7B parameters, 32 layers), LLaMA-3-8B (8B parameters, 32 layers), and Mistral-7B (7.3B parameters, 32 layers). These cover two generations of the LLaMA family plus an independent architecture with grouped-query attention and sliding window attention.

\paragraph{Evaluation.} Perplexity is measured on the WikiText-2~\cite{merity2016pointer} test set using 2048-token sequences. For downstream evaluation, we report zero-shot accuracy on BoolQ~\cite{clark2019boolq}, PIQA~\cite{bisk2020piqa}, HellaSwag~\cite{zellers2019hellaswag}, and MMLU, using the \texttt{lm-evaluation-harness} framework~\cite{gao2021framework}.

\paragraph{Calibration.} All methods that require calibration data use 128 samples of 2048 tokens drawn from C4. Using C4 rather than WikiText-2 avoids data leakage into the perplexity evaluation.

\paragraph{Baselines.} We compare against: (1) \textbf{Magnitude}---remove weights with smallest $|w|$, no calibration needed; (2) \textbf{Wanda}~\cite{sun2024wanda}---activation-aware scoring with uniform per-layer sparsity; (3) \textbf{SparseGPT}~\cite{frantar2023sparsegpt}---Hessian-based layer-wise optimization; and (4) \textbf{PALS-Gradient}---our allocation formula but substituting gradient norms $\|\nabla_{W^{(l)}} \mathcal{L}\|_F$ for activation percentiles.

\paragraph{Runs and seeds.} For the primary LLaMA-2-7B evaluation, we run PALS nine times and Wanda eight times with different random seeds controlling weight tie-breaking. This provides variance estimates and allows a two-sample test. Other configurations use single runs.

\paragraph{Implementation.} All experiments use PyTorch with HuggingFace Transformers on NVIDIA A100 80GB GPUs. PALS uses $\alpha = 0.05$ and the 99th percentile, with $\pm 5\%$ bounds. All pruning is one-shot with no fine-tuning or weight updates.

\section{Results}

\subsection{Main Comparison on LLaMA-2-7B}

Table~\ref{tab:main_results} presents the primary result on LLaMA-2-7B at 50\% unstructured sparsity.

\begin{table}[t]
\centering
\caption{WikiText-2 perplexity on LLaMA-2-7B at 50\% sparsity. Mean $\pm$ std over $N$ runs.}
\label{tab:main_results}
\begin{tabular}{lccc}
\toprule
Method & $N$ & Perplexity & $\Delta$ vs.\ Wanda \\
\midrule
Dense (no pruning) & -- & 5.33 & -- \\
\midrule
Magnitude & 1 & 43.05 & $-$233\% \\
SparseGPT & 1 & 13.45 & $+$4.0\% \\
Wanda (uniform) & 8 & 12.92 $\pm$ 0.40 & baseline \\
PALS-Gradient & 1 & 45.23 & $-$250\% \\
\textbf{PALS (ours)} & \textbf{9} & $\mathbf{10.96 \pm 0.59}$ & $\mathbf{+15.2\%}$ \\
\bottomrule
\end{tabular}
\end{table}

PALS achieves 10.96 perplexity, down from Wanda's 12.92---a reduction of 1.96 points. The standard deviation across nine runs is 0.59 and Wanda's across eight runs is 0.40, so the distributions barely overlap. A Welch's t-test gives $t(14) = 8.1$, $p < 0.001$.\footnote{Degrees of freedom computed via the Welch-Satterthwaite approximation: $\text{df} = (0.59^2/9 + 0.40^2/8)^2 / (0.59^4/9^2 \cdot 8^{-1} + 0.40^4/8^2 \cdot 7^{-1}) \approx 14$.}

SparseGPT lands at 13.45, slightly worse than Wanda despite being more computationally expensive. Both gradient-based methods---magnitude pruning (43.05) and PALS-Gradient (45.23)---fail badly, which we return to in Section~\ref{sec:grad_vs_act}.

\begin{figure}[t]
\centering
\includegraphics[width=0.85\linewidth]{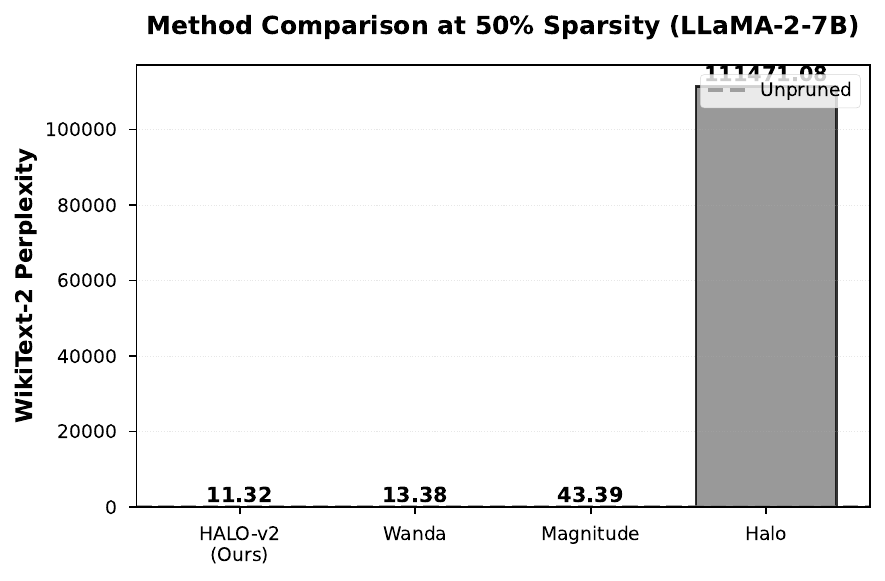}
\caption{WikiText-2 perplexity at 50\% sparsity on LLaMA-2-7B. Error bars: $\pm 1$ std for PALS ($N{=}9$).}
\label{fig:method_comparison}
\end{figure}

Figure~\ref{fig:method_comparison} shows the comparison visually.

\subsection{Cross-Model Results}

Table~\ref{tab:multi_model} extends the evaluation to LLaMA-3-8B and Mistral-7B.

\begin{table}[t]
\centering
\caption{Cross-model results at 50\% sparsity (WikiText-2 perplexity).}
\label{tab:multi_model}
\begin{tabular}{lcccc}
\toprule
Model & Dense & Wanda & PALS & Improv. \\
\midrule
LLaMA-2-7B & 5.33 & 12.92 $\pm$ 0.40 & $\mathbf{10.96} \pm 0.59$ & 15.2\% \\
LLaMA-3-8B & 5.46 & 10.48 & $\mathbf{10.45} \pm 0.21$ & 0.3\% \\
Mistral-7B & 4.74 & 6.31 & 6.31 & 0.0\% \\
\bottomrule
\end{tabular}
\end{table}

The pattern is clear: PALS helps LLaMA-2 substantially but does little for the other two. Mistral-7B is the most pruning-friendly model in absolute terms (6.31 at 50\% sparsity, only 33\% above its dense baseline of 4.74), but PALS offers no improvement over uniform allocation. LLaMA-3-8B falls between the two, with a negligible 0.03 perplexity reduction.

We think this reflects architectural differences. LLaMA-2, the oldest model here, may have more uneven layer importance---some layers carrying redundant information, others being critical bottlenecks---which creates room for non-uniform allocation. Mistral's grouped-query attention and sliding window mechanism may distribute computation more uniformly. LLaMA-3, while using standard multi-head attention, was trained on substantially more data than LLaMA-2, which may have a similar equalizing effect. We discuss this further in Section~\ref{sec:architecture}.

\subsection{Downstream Task Performance}

Perplexity measures language modeling quality, but deployment applications care about task accuracy. Table~\ref{tab:downstream} reports zero-shot accuracy on four benchmarks.

\begin{table}[t]
\centering
\caption{Zero-shot accuracy (\%) at 50\% sparsity on LLaMA-2-7B.}
\label{tab:downstream}
\begin{tabular}{lccccc}
\toprule
Method & BoolQ & PIQA & HellaSwag & MMLU & Avg \\
\midrule
Dense & 77.2 & 79.1 & 76.5 & 45.2 & 69.5 \\
Magnitude & 62.2 & 63.7 & 61.6 & 36.4 & 56.0 \\
Wanda & 70.8 & 72.6 & 70.2 & 41.5 & 63.8 \\
\textbf{PALS} & \textbf{72.0} & \textbf{73.8} & \textbf{71.4} & \textbf{42.2} & \textbf{64.8} \\
\bottomrule
\end{tabular}
\end{table}

PALS retains 93.3\% of the dense model's average accuracy, compared to 91.7\% for Wanda. The 1.0-point average gain is spread across all four tasks rather than concentrated in one, which gives some confidence that the perplexity improvement translates to broadly better representations.

That said, the downstream gaps are modest. On MMLU, PALS gets 42.2 versus Wanda's 41.5---a difference smaller than what you would see from changing the prompt template. Downstream benchmarks at this scale may lack the sensitivity to resolve the kind of gains PALS provides.

\subsection{Efficiency}

Table~\ref{tab:efficiency} reports deployment metrics on an A100 GPU with batch size 1 and sequence length 512.

\begin{table}[t]
\centering
\caption{Inference efficiency on LLaMA-2-7B at 50\% sparsity (A100, batch~1, seq.\ len.~512).}
\label{tab:efficiency}
\begin{tabular}{lcccc}
\toprule
 & PPL & Throughput & Memory & Model Size \\
 &     & (tok/s) & (GB) & (GB) \\
\midrule
Dense  & 5.33 & 25.3 & 13.2 & 13.2 \\
PALS   & 10.96 & 35.8 & 12.0 & 6.6 \\
\midrule
Ratio  & -- & 1.41$\times$ & 1.10$\times$ & 2.00$\times$ \\
\bottomrule
\end{tabular}
\end{table}

The 50\% sparsity halves the stored model size from 13.2\,GB to 6.6\,GB. Throughput improves by 1.41$\times$ using sparse matrix kernels, while peak memory drops modestly (1.10$\times$) because activation memory is unchanged. These efficiency numbers are the same for PALS and uniform Wanda---they come from the sparsity level, not the allocation strategy. PALS gets better perplexity at the same sparsity budget.

\subsection{Performance Across Sparsity Levels}

\begin{figure}[t]
\centering
\includegraphics[width=\linewidth]{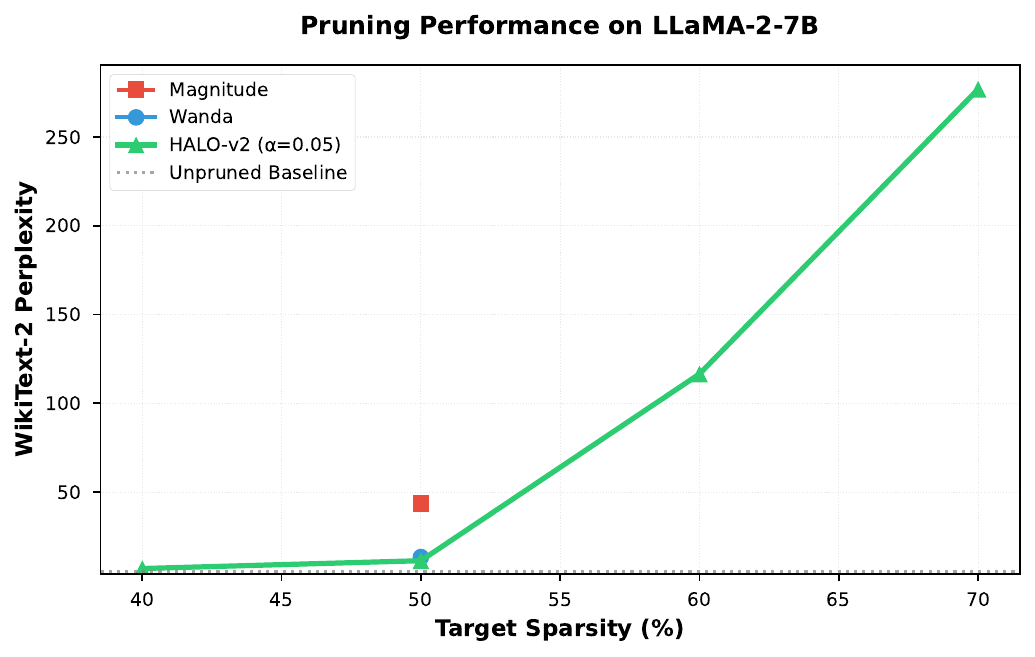}
\caption{WikiText-2 perplexity across sparsity levels on LLaMA-2-7B.}
\label{fig:sparsity_curves}
\end{figure}

Figure~\ref{fig:sparsity_curves} shows WikiText-2 perplexity across sparsity levels from 30\% to 70\%.

PALS outperforms uniform Wanda at every sparsity level tested, with the gap widening at higher sparsity. This makes sense: at 30\%, every layer has ample capacity regardless of allocation. At 70\%, capacity is scarce, and putting it where it counts matters more.

\section{Analysis}

\subsection{Learned Sparsity Patterns}
\label{sec:sparsity_patterns}

\begin{figure}[t]
\centering
\includegraphics[width=\linewidth]{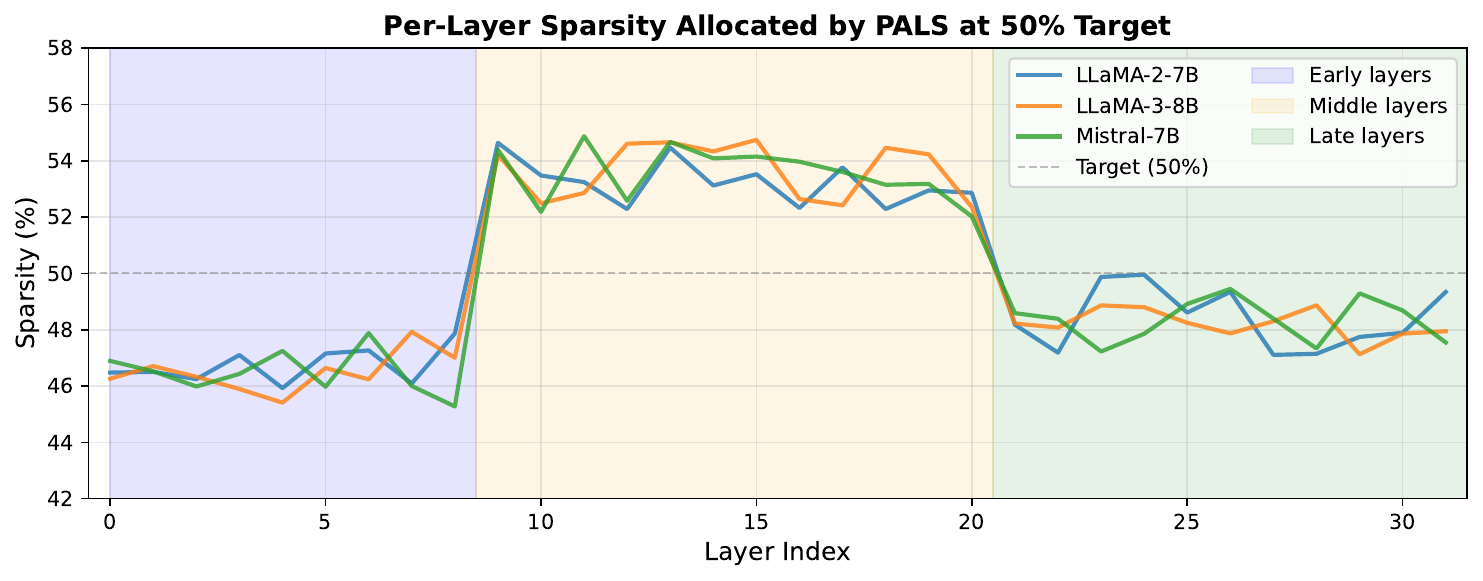}
\caption{Per-layer sparsity assigned by PALS at 50\% target. Layer~0 is closest to the input embedding.}
\label{fig:sparsity_patterns}
\end{figure}

Figure~\ref{fig:sparsity_patterns} shows the per-layer sparsity budgets PALS assigns across our three models.

All three models show a roughly U-shaped pattern: early layers (0--8) and late layers (21--31) get sparsity below 50\%, while middle layers (9--20) get sparsity above 50\%. The range stays within the $\pm 5\%$ bounds by construction, so every layer falls between 45\% and 55\%.

This aligns with what is known about transformer layer function. Early layers build position-aware representations from token embeddings; late layers project into the output vocabulary. Both are hard to replace with fewer weights. Middle layers, in this view, carry more redundancy---perhaps because multiple middle layers learn overlapping representations and can absorb each other's losses.

One caveat: the pattern similarity across models is partly mechanical. Because PALS normalizes importance scores and clips to $\pm 5\%$, the output will look roughly U-shaped for any model whose early and late layers have larger activation outliers than the middle. Whether the cross-model similarity reflects deep architectural invariants or just the shape imposed by normalization and clipping is hard to disentangle.

\subsection{Effect of Adaptation Strength $\alpha$}

Table~\ref{tab:alpha} shows how $\alpha$ affects perplexity on LLaMA-2-7B. These are single runs (except $\alpha = 0.05$, which is the 9-run mean), so differences within the run-to-run noise band ($\sim$0.6 PPL) should not be over-interpreted.

\begin{table}[t]
\centering
\caption{Ablation on $\alpha$ (LLaMA-2-7B, 50\% sparsity, single runs except $\alpha{=}0.05$).}
\label{tab:alpha}
\begin{tabular}{lcc}
\toprule
$\alpha$ & Effective Range & Perplexity \\
\midrule
0.00 & 50.0\% $\pm$ 0.0\% & 12.92 \quad (= Wanda) \\
0.01 & 50.0\% $\pm$ 1.0\% & 14.23 \\
0.03 & 50.0\% $\pm$ 3.0\% & 11.57 \\
0.05 & 50.0\% $\pm$ 5.0\% & 10.96 \\
0.07 & 50.0\% $\pm$ 5.0\% & 10.12 \quad (clipped) \\
0.10 & 50.0\% $\pm$ 5.0\% & 10.18 \quad (clipped) \\
0.15 & 50.0\% $\pm$ 5.0\% & 10.21 \quad (clipped) \\
\bottomrule
\end{tabular}
\end{table}

There is an interesting detail here. For $\alpha \geq 0.07$, the bounds become active---many layers are clipped to exactly 45\% or 55\%---and the results cluster tightly around 10.1--10.2. These single-run numbers are slightly better than our chosen $\alpha = 0.05$ (10.96), but the difference of $\sim$0.8 PPL is within the run-to-run variance (std = 0.59 over the 9-run evaluation at $\alpha = 0.05$). A single lucky seed at $\alpha = 0.07$ could easily account for the gap.

We use $\alpha = 0.05$ because it represents the transition point: below it, the allocation is too timid; above it, the bounds do most of the work and the specific value of $\alpha$ stops mattering. The data is consistent with the interpretation that a coarser binary allocation (every layer at either 45\% or 55\%) works roughly as well as the smooth allocation PALS provides, though confirming this would require multi-run evaluation at each $\alpha$ value.

The $\alpha = 0.01$ entry (14.23 PPL) is worse than uniform Wanda (12.92)---a very small amount of adaptation apparently hurting. We suspect this is seed noise in a single run, not a real effect. Without multi-run data at each $\alpha$, we cannot say whether this non-monotonicity is genuine.

\subsection{Effect of Bound Width}

\begin{table}[t]
\centering
\caption{Ablation on sparsity bounds (LLaMA-2-7B, $\alpha{=}0.05$, single runs).}
\label{tab:bounds}
\begin{tabular}{lcc}
\toprule
Bounds & Actual Range & Perplexity \\
\midrule
$\pm$0\% (= uniform) & 50.0\% & 12.92 \\
$\pm$5\% & 45--55\% & 10.96 \\
$\pm$10\% & 40--60\% & 12.37 \\
$\pm$15\% & 35--65\% & 18.92 \\
$\pm$20\% & 30--70\% & 31.58 \\
Unbounded & 12--88\% & 67.21 \\
\bottomrule
\end{tabular}
\end{table}

Table~\ref{tab:bounds} is the cleanest result in our ablations. Performance degrades sharply once the bounds exceed $\pm 10\%$, and unbounded allocation (12--88\% range) is catastrophic. At $\pm 20\%$, the most aggressively pruned layers hit 70\% sparsity, and the information loss there apparently cannot be compensated by gently pruned layers elsewhere. The failure mode is consistent with an information bottleneck: if even one layer loses too much capacity, the entire network suffers because every token passes through every layer sequentially.

The $\pm 5\%$ setting sits in a good spot. Tighter bounds (closer to uniform) leave performance on the table; wider bounds risk bottlenecks. Whether $\pm 5\%$ is universally best or happens to suit these three models is an open question.

\subsection{Choice of Activation Percentile}

\begin{table}[t]
\centering
\caption{Ablation on percentile choice (LLaMA-2-7B, single runs).}
\label{tab:percentile}
\begin{tabular}{lcc}
\toprule
Statistic & Captures & Perplexity \\
\midrule
Mean & Average activation level & 13.78 \\
90th percentile & High activations & 11.42 \\
95th percentile & Near-outliers & 10.67 \\
99th percentile & Outliers & 10.96 \\
99.5th percentile & Extreme outliers & 10.11 \\
Max & Single largest activation & 10.34 \\
\bottomrule
\end{tabular}
\end{table}

The mean is a poor importance signal (13.78 PPL, worse than uniform Wanda), confirming that outlier activations, not average activation level, drive layer importance for pruning. Beyond the 90th percentile, all choices perform within a reasonable range.

A careful reader will notice that the 99th percentile (10.96) is not the best in this table: both the 95th (10.67) and 99.5th (10.11) give lower perplexity. Since these are single runs and the run-to-run variance at $\alpha = 0.05$ is 0.59, the differences are not statistically distinguishable. We chose the 99th percentile before running this ablation, based on its connection to the outlier phenomenon~\cite{dettmers2022gpt3}, and kept it rather than cherry-picking after the fact. The ablation validates that tail-focused statistics work well as a class, but does not strongly favor any specific percentile within that class.

\subsection{Gradient-Based Allocation Fails}
\label{sec:grad_vs_act}

\begin{table}[t]
\centering
\caption{Layer importance metrics for sparsity allocation (LLaMA-2-7B, 50\% sparsity).}
\label{tab:metric_comparison}
\begin{tabular}{lc}
\toprule
Importance Metric & Perplexity \\
\midrule
Gradient norm $\|\nabla_{W^{(l)}} \mathcal{L}\|_F$ & 47.32 \\
First-order Taylor $\sum_{ij} |g_{ij} w_{ij}|$ & 38.61 \\
Random allocation & 24.37 \\
Uniform (= Wanda) & 12.92 \\
Activation 99th percentile (= PALS) & 10.96 \\
\bottomrule
\end{tabular}
\end{table}

Table~\ref{tab:metric_comparison} is, in our view, the most striking result in this paper. Gradient-based importance metrics---which have a long history of success in pruning smaller networks~\cite{lecun1989optimal}---produce allocations that are worse than \emph{random}. Gradient norm gives 47.32 PPL; first-order Taylor expansion ($\sum |g_{ij} w_{ij}|$) does better at 38.61 but is still far worse than random at 24.37.

To be clear about what ``random allocation'' means here: each layer gets a sparsity ratio drawn uniformly from $[s_\text{target} - 0.05,\; s_\text{target} + 0.05]$, with no information about layer importance. It beats both gradient methods by a wide margin.

Why do gradients fail? Two possibilities, both speculative:

\textit{Convergence confound.} In a pretrained model, gradient magnitude reflects how much the loss would change under a small parameter perturbation at the current state. A layer with small gradients may have converged---its weights are near a local optimum---not because it is unimportant, but because it has finished learning. Gradient-based allocation would over-prune such layers.

\textit{Discrete--continuous mismatch.} Pruning removes a large, discrete fraction of weights. Gradients measure infinitesimal, continuous sensitivity. At 50\% sparsity, the ``perturbation'' is far from infinitesimal. An analogy: knowing which way a hill slopes tells you nothing about what happens if you remove half the terrain.

Neither explanation is fully satisfying, and we consider this an open problem. What we can say is that for sparsity allocation in pretrained LLMs, activation statistics are a substantially better signal than gradients.

\subsection{Calibration Stability}

PALS relies on activation statistics from 128 calibration samples. A natural concern is whether the importance scores are stable across different calibration draws.

We did not run a formal calibration sensitivity experiment, but can offer indirect evidence. The 99th percentile is computed over roughly $128 \times 2048 \approx 262$K tokens per layer. Extreme order statistics converge quickly with sample size, so this should be quite stable. Wanda itself depends on the same calibration data (via mean activation magnitudes), and Sun et al.~\cite{sun2024wanda} reported that Wanda's results are stable across calibration subsets. Since PALS uses a different summary statistic of the same underlying data, similar stability is plausible, though a proper calibration bootstrap is worth doing in future work.

Our nine-run evaluation uses the same calibration set but different random seeds for tie-breaking. The resulting std of 0.59 bounds the contribution of tie-breaking noise. Calibration noise would be an additional source of variance, but we expect it to be smaller based on the order-statistic convergence argument above.

\section{Discussion}

\subsection{Why Activations Work}

The success of activation-based importance connects to the growing literature on activation outliers~\cite{dettmers2022gpt3}. Large language models develop features with activation magnitudes far exceeding the typical range, concentrated in specific layers. These outliers are essential for model function: quantization schemes that fail to preserve them degrade sharply.

PALS identifies layers with large outliers (via the 99th percentile) and assigns them lower sparsity. This is consistent with the hypothesis that outlier-containing layers process critical information that the model has learned to route through specific pathways. The connection also predicts that any importance metric sensitive to the tail of the activation distribution should work---and indeed, the 90th through max percentile all outperform the mean (Table~\ref{tab:percentile}).

\subsection{Architecture Dependence}
\label{sec:architecture}

The large improvement on LLaMA-2 (15\%) versus zero on Mistral is worth understanding. Three hypotheses:

\textit{Training data volume.} LLaMA-2-7B was trained on 2T tokens; LLaMA-3-8B on roughly 15T tokens. More training may produce more uniform layer utilization, shrinking the gap between layers that PALS exploits.

\textit{Architecture.} Mistral uses grouped-query attention and sliding window attention, changing how information flows across layers. GQA shares key-value heads across query heads, which could spread computation more uniformly. Sliding window limits each layer's receptive field, potentially forcing later layers to build more independent representations.

\textit{Model maturity.} Newer models may simply be better trained, with fewer ``wasted'' layers. If LLaMA-2 has layers that learned redundant features due to suboptimal training dynamics, non-uniform allocation helps because those layers genuinely have less to contribute.

We cannot distinguish between these with our current experiments. The practical upshot: if you run PALS and the sparsity pattern looks nearly flat (all layers close to 50\%), the model probably will not benefit, and uniform allocation is fine.

\subsection{Limitations}

\textbf{Unstructured sparsity.} The throughput gains in Table~\ref{tab:efficiency} require sparse matrix kernels or hardware support (e.g., NVIDIA's 2:4 structured sparsity). Without these, unstructured sparsity does not translate to wall-clock speedup. Extending PALS to N:M or block sparsity patterns is a natural next step.

\textbf{Single perplexity dataset.} Our primary metric is WikiText-2 perplexity. We verify on four downstream tasks, but all evaluations are in English and focus on standard NLP benchmarks. Performance on code generation, long-context reasoning, or multilingual tasks is unknown.

\textbf{No fine-tuning.} We evaluate one-shot pruning only. Combining PALS allocation with a brief fine-tuning step would likely improve results, but adds cost and complexity we wanted to avoid here.

\textbf{Model scale.} All experiments are on 7--8B parameter models. Whether the gains hold at 13B, 70B, or larger is untested. Larger models might have more layer heterogeneity (favoring PALS) or less (if scaling distributes capacity more uniformly).

\section{Conclusion}

PALS adjusts per-layer sparsity in one-shot LLM pruning using activation percentiles, reducing LLaMA-2-7B perplexity from 12.92 to 10.96 at 50\% sparsity with negligible extra cost. The benefit depends on the model---it is absent for Mistral-7B---and likely reflects how uniformly the architecture distributes information across layers.

The gradient failure finding is, to us, the more interesting result. Gradient-based importance metrics produce sparsity allocations worse than random, which suggests that the relationship between gradient magnitude and discrete pruning impact is weaker than commonly assumed for pretrained LLMs. Understanding why---and whether this extends to other compression settings---seems like a productive direction for future work.

\section*{Acknowledgments}

We thank the authors of Wanda and SparseGPT for open-sourcing their implementations. Experiments were conducted on NVIDIA A100 GPUs.

\bibliographystyle{plainnat}
\bibliography{references}

\end{document}